\documentclass[runningheads]{llncs}

\usepackage[T1]{fontenc}
\usepackage{graphicx}
\usepackage{booktabs}
\usepackage{amsmath,amssymb}
\usepackage{tikz}
\usetikzlibrary{arrows.meta,positioning,fit,backgrounds}
\usepackage{hyperref}
\usepackage{url}
\usepackage{xcolor}
\usepackage[misc]{ifsym}
\usepackage[ruled,lined,noend]{algorithm2e}

\begin{document}

\title{Recovering Business Rules as Cascade Lookups for Enterprise Field Prediction}
\titlerunning{Business Rule Cascades for Field Prediction}
\toctitle{Recovering Business Rules as Cascade Lookups for Enterprise Field Prediction}

\author{Changxu Duan~(\Letter)\orcidID{0000-0003-0547-0901}}
\authorrunning{C. Duan}
\tocauthor{Changxu Duan}
\institute{Technische Universit{\"a}t Darmstadt\\ Darmstadt, Germany\\
\email{duan@linglit.tu-darmstadt.de}}

\maketitle

\begin{abstract}
Enterprise sales-order systems rely on manually maintained configuration rules to determine fields such as payment terms, shipping points, and sales ownership. When these rules are incomplete or stale, operators must enter values by hand, and downstream errors spread into billing, logistics, and reporting. This paper studies whether missing or outdated rules can be recovered from historical transactions instead of being relearned as opaque statistical patterns. We introduce a knowledge-graph-guided workflow in which an LLM agent proposes candidate business keys, meaning column combinations such as customer and document type, validates them through SQL profiling, and compiles the resulting evidence into deterministic cascade lookups. The LLM is used only during offline rule discovery. At inference time, prediction is a transparent key lookup with no model-serving requirement. Experiments on enterprise sales-order prediction and public relational benchmarks show that cascades are most effective when fields follow stable business rules, while learned graph models remain preferable when prediction depends on temporal or neighborhood patterns. Drift analysis further turns residual errors into diagnostics for configuration maintenance.

\end{abstract}

\noindent\textbf{Code and data.} \url{https://github.com/Fireblossom/CascadER}

\section{Introduction}
\label{sec:intro}

Enterprise Resource Planning (ERP) systems orchestrate business operations, spanning procurement, manufacturing, sales, and finance, by maintaining a unified transactional database across organizational functions. In a sales-order workflow, the system must fill business fields before an order can be billed, shipped, and reported. These fields include payment terms that determine invoice due dates, shipping points that identify the warehouse or dispatch location, Incoterms that allocate shipping cost and risk, and sales ownership fields that assign the responsible sales team. These values are not arbitrary labels. In well-configured ERP systems, they are determined by business rules over customers, materials, document types, and organizational units such as sales offices, sales groups, and plants.

SAP, the dominant ERP vendor serving over 400{,}000 customers worldwide, implements such field-level determination through its Condition Technique, SAP's rule mechanism for assigning default field values from ordered lookup tables called Access Sequences~\cite{klein2024salt,hoffart2025fmslt}. Each table is keyed by a combination of business attributes such as customer, material, and organizational context. However, these Access Sequences must be manually configured and maintained by domain experts. When configurations are incomplete, outdated after system migrations, or invalidated by organizational restructuring, the system either leaves fields blank or applies coarse defaults, forcing sales representatives to select values manually~\cite{haug2011classification}.

This configuration gap has tangible business consequences. An incorrect shipping point routes goods from the wrong warehouse and causes fulfillment delays. Mismatched payment terms create billing disputes. Wrong Incoterms misallocate shipping risk between buyer and seller, and inaccurate sales group assignments distort commission calculations. Systematic reviews of enterprise and manufacturing data ecosystems confirm that such data quality defects are a persistent and costly problem~\cite{cui2020manufacturing,riahi2021artificial}. Furthermore, these errors are not static because even well-configured rules degrade over time as organizations restructure and customer relationships evolve. On the SALT-KG benchmark~\cite{mulang2025saltkg}, which captures 2.3M historical order items from a single SAP S/4HANA ERP system instance, between 17\% and 44\% of customers changed their most frequently assigned field value within a two-year window, depending on the field (Section~\ref{sec:drift}). Because each affected customer generates many orders, stale defaults can propagate errors across thousands of transactions before anyone notices.

A prediction system that suggests correct field values during order entry would reduce the need for manual correction, catch configuration drift before it propagates, and provide configuration owners with continuous visibility into which rules are still valid and which have silently broken. Here, configuration drift means that a formerly valid default no longer matches current business assignments because contracts, organizational responsibilities, or logistics settings have changed. We focus on cascade lookups as a practical representation for such suggestions. A cascade lookup is an ordered list of lookup tables, from specific rules to general fallbacks. For example, the system may first try a customer-by-document-type rule, then a customer-only rule, and finally a global default.

Existing approaches treat field prediction as a standard supervised learning problem, but each family of methods encounters a characteristic barrier when applied to rule-governed enterprise data. Gradient-boosted ensembles~\cite{chen2016xgboost,prokhorenkova2018catboost,erickson2020autogluon} perform well on fields with few possible values but struggle on high-cardinality targets whose correct value depends on specific multi-column business rules rather than broad feature correlations. For example, on the SALT-KG benchmark, predicting which internal sales team handles an account (589 possible teams) requires knowing the exact combination of customer and organizational unit that the company's assignment policy prescribes, and the best tree-based model achieves only 51\% MRR on this field. Heterogeneous GNNs~\cite{hamilton2017graphsage,robinson2024relbench} learn relational embeddings by propagating signals along primary-foreign key edges, yet they too fail on fields where the correct value is determined by a specific lookup rule rather than by neighborhood similarity. On SALT-KG, a GNN achieves only 15.8\% accuracy on sales team assignment and 37.5\% on customer payment terms. KG-enhanced embeddings~\cite{mulang2025saltkg} inject domain metadata as auxiliary features, but yield only modest accuracy gains ($<$2 percentage points) when limited to field definitions without encoding the determination rules themselves. The common limitation is that these methods model the problem as statistical pattern recognition, while the underlying signal is often configured determinism. By configured determinism, we mean that field values follow explicit lookup rules keyed by specific combinations of customer, document type, material, and organizational unit.

Directly using LLMs at inference time is also impractical. Even with retrieval-augmented approaches, per-order LLM calls introduce latency and cost incompatible with high-throughput transactional systems, and enterprise deployments require that identical inputs always produce identical outputs with full audit trails, a guarantee that stochastic generation cannot provide.

Our key observation is that configured rules, even when missing from the live system, leave recoverable traces in historical transaction data. The challenge is that the space of possible key combinations is combinatorially large, making exhaustive search difficult, while identifying which combinations carry business meaning requires domain knowledge that is not encoded in table schemas alone. We address this by using an LLM agent guided by a Knowledge Graph~\cite{mulang2025saltkg} that provides field-level descriptions and dependency hints derived from SAP's internal ontology. The KG helps prioritize semantically meaningful key combinations, while SQL profiling against the full transaction history validates each candidate rule with coverage and accuracy statistics. The output is a set of deterministic JSON lookup cascades that require no GPU or LLM at inference time, provide built-in interpretability, and can be rebuilt as configurations change.

This paper makes three contributions.
\begin{enumerate}
    \item A KG-guided rule discovery workflow in which an LLM proposes candidate business keys, profiles them via SQL, and codifies the selected rules as hierarchical cascade lookups, achieving the best results among evaluated baselines on all 8 SALT-KG fields.
    \item A generalization study on two public RelBench~v2 datasets (rel-arxiv, rel-stack) that characterizes when cascades match or underperform GNNs, showing that the method's advantage is specific to domains with stable, recoverable rules.
    \item A drift root-cause analysis linking many residual errors to organizational restructuring and showing how cascade failures can support configuration maintenance.
\end{enumerate}

\paragraph{Application setting.}
An SAP sales order consists of a header and one or more line items. The header stores order-wide attributes, while each line item stores product-specific attributes for one ordered product. The 8~target fields in SALT-KG span both levels (Table~\ref{tab:fields}). Our system targets cases where configured Access Sequences fail to resolve a value. The cascade lookup provides an editable suggestion based on statistically recovered rules, with cascade-level provenance visible to the operator.

\begin{table}[t]
\centering
\caption{Target field classification. Header-level fields apply to the entire order. Item-level fields are set per product line. Downstream consequence describes the operational impact of an incorrect value.}
\label{tab:fields}
\small
\begin{tabular}{llp{5.8cm}}
\toprule
Field & Level & Downstream consequence of error \\
\midrule
\textsc{CustomerPaymentTerms} & Header & Incorrect invoice due dates and early/late payment disputes \\
\textsc{SalesGroup} & Header & Misattributed revenue reporting and incorrect commission assignment \\
\textsc{SalesOffice} & Header & Distorted regional sales analytics \\
\textsc{HeaderIncoterms} & Header & Wrong allocation of shipping risk and cost between buyer and seller \\
\textsc{ShippingCondition} & Header & Suboptimal carrier selection and delivery delays \\
\textsc{ItemIncoterms} & Item & Per-line risk allocation inconsistent with contract \\
\textsc{ShippingPoint} & Item & Goods dispatched from wrong warehouse and fulfillment delay \\
\textsc{Plant} & Item & Wrong production or sourcing site and inventory discrepancy \\
\bottomrule
\end{tabular}
\end{table}

\section{Related Work}
\label{sec:related}

\paragraph{Relational deep learning.}
RelBench~\cite{robinson2024relbench} established standardized benchmarks for prediction on multi-table relational databases, proposing heterogeneous GNN baselines that construct message-passing graphs from primary and foreign key relationships. RelBench~v2~\cite{gu2026relbenchv2} extends this with autocomplete tasks, temporal splits, and stronger GNN/LightGBM baselines. Our method differs fundamentally because instead of learning node embeddings via gradient descent, we generate deterministic lookup tables guided by domain knowledge.

\paragraph{Knowledge graphs for structured prediction.}
KGs have been used to inject domain semantics into ML pipelines, e.g., as entity embeddings~\cite{wang2021kepler}, relation-aware features, or constraint layers. Recent visions increasingly emphasize unifying KGs with LLMs to enable synergistic reasoning~\cite{pan2024unifying}. In enterprise settings, ontologies can describe business concepts and determination dependencies~\cite{hoffart2025fmslt}. Our approach uses the KG as a search guide rather than as an embedding source. The LLM uses KG-provided descriptions to propose SQL profiling candidates, while accepted rules are selected by deterministic validation.

\paragraph{LLM-based code generation for data tasks.}
Recent work applies LLMs as autonomous data analysts, generating SQL queries~\cite{pourreza2024dinsql,gao2023texttosql}, reasoning directly over tabular data~\cite{zhang2025tablellm}, writing feature engineering code, and performing iterative data exploration~\cite{hong2024data}. ReAct~\cite{yao2023react} demonstrated that interleaving reasoning traces with actions improves task grounding. Our method builds on this paradigm but targets a specific outcome, namely the systematic discovery and codification of deterministic business rules into deployable prediction cascades rather than general-purpose analysis.

\paragraph{Tabular foundation models.}
TabPFN~\cite{hollmann2023tabpfn} introduced in-context learning for tabular prediction, achieving strong results on small-to-medium datasets without task-specific training. ConTextTab~\cite{spinaci2025contexttab} extends this paradigm with semantic embeddings from pretrained language models. However, both architectures face scalability constraints that preclude direct application to enterprise transaction data. TabPFN's in-context mechanism is designed for datasets with up to tens of thousands of samples, while ConTextTab requires 80\,GB GPU memory with a context window of 8{,}192 rows and bagging for larger tables. SALT-KG's 2.3M rows and targets with up to 589 classes fall outside these operating regimes. Our cascade approach sidesteps these limitations. Rule discovery uses an LLM offline, and the resulting JSON lookups require no model serving at inference time.

\paragraph{Enterprise field prediction.}
The SALT benchmark~\cite{klein2024salt} introduced the task of predicting SAP sales order fields with tree-based and GNN baselines. SALT-KG~\cite{mulang2025saltkg} extended it with a metadata Knowledge Graph extracted from SAP's internal ontology via curated SPARQL queries, showing that KG-derived embeddings yield modest MRR improvements. The FMSLT vision~\cite{hoffart2025fmslt} argues that enterprise prediction should integrate declarative knowledge, such as ontologies, with procedural knowledge, such as determination procedures. Our approach operationalizes part of the declarative layer through KG-guided SQL profiling, while the drift analysis shows where transaction history alone cannot capture later procedural or organizational changes.

\section{Method}
\label{sec:method}

\subsection{Problem Formulation}
\label{sec:problem}

We consider a temporal relational autocomplete task. For a target field~$f$, each historical transaction~$i$ contains an input tuple~$x_i$, a timestamp~$t_i$, and a target value~$y_i \in \mathcal{Y}_f$. Training records satisfy $t_i < T$, and evaluation records satisfy $t_i \geq T$, where $T$ is the temporal cutoff. The task is to predict~$y_i$ from the fields available at order-entry time.

A composite key $K=(c_1,\ldots,c_d)$ selects a small set of input columns. For a record~$i$, $x_i[K]$ denotes the corresponding key value. A lookup table for~$K$ maps each observed key value~$k$ to target values ranked by their empirical frequency among training records with $x_i[K]=k$, subject to a minimum support threshold.

A cascade is an ordered list of lookup tables, from more specific keys to more general fallbacks. At inference time, the system first tries the most specific key. If no matching entry exists, it falls back to the next level, and eventually to the global target distribution. The objective is to choose a short cascade that gives accurate, high-coverage, temporally valid, and auditable predictions.

\begin{figure}[t]
\centering
\includegraphics[width=\linewidth]{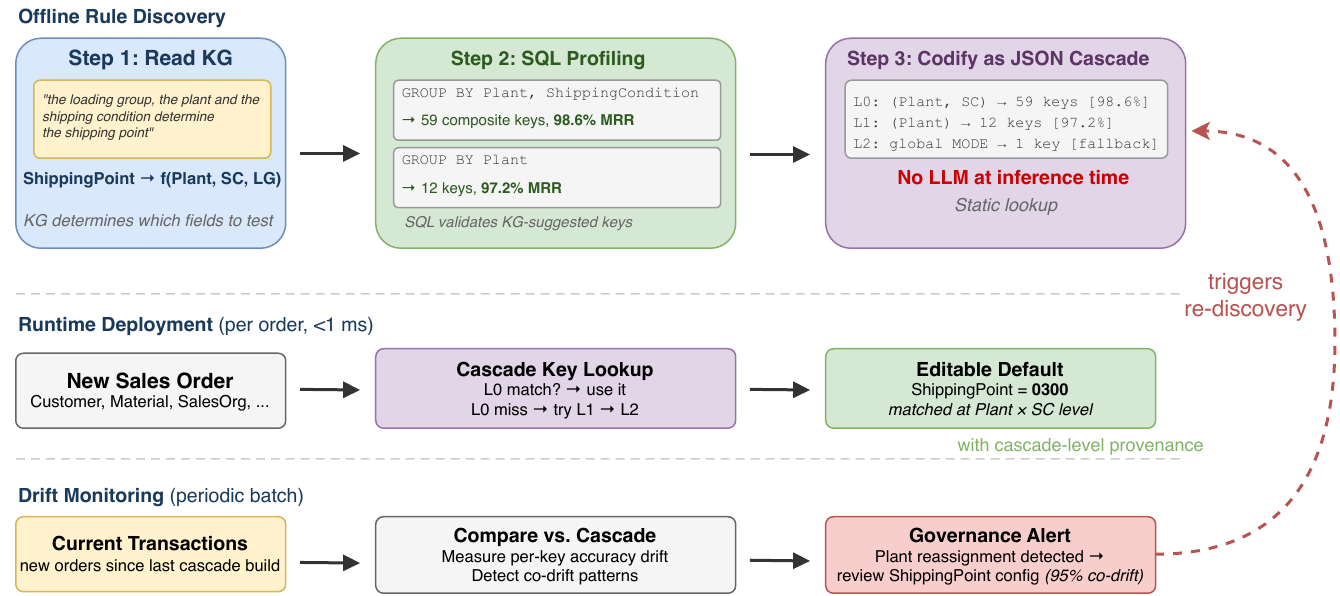}
\caption{System overview. The top panel shows offline rule discovery producing JSON cascade files. The middle panel shows runtime key lookup providing editable suggestions during order entry. The bottom panel shows drift monitoring comparing current transactions against cascade mappings and surfacing alerts for governance review.}
\label{fig:system}
\end{figure}

Figure~\ref{fig:system} provides an overview of the end-to-end workflow, from offline rule discovery through runtime deployment.

\subsection{Rule Discovery (Explore)}
\label{sec:discovery}

For each target field~$f$, the LLM operates as a rule discovery agent in an iterative propose, profile, score, and refine loop inspired by ReAct~\cite{yao2023react}. The agent consumes, as a read-only input, the metadata Knowledge Graph published by SALT-KG~\cite{mulang2025saltkg}, which was constructed by the SALT-KG authors through curated SPARQL queries against SAP's internal ontology. These queries are used as published and were not authored or modified for this work. At each iteration the agent maintains a state consisting of (i)~semantic hints extracted from the KG's natural-language descriptions of~$f$ and its related entities, (ii)~a set of candidate composite keys under consideration, (iii)~SQL profiling evidence such as coverage, purity, and support counts, and (iv)~validation accuracy and residual error buckets for the current cascade. SQL profiling refers to grouped aggregate queries that estimate how often a candidate key is observed and how consistently it maps to a single target value.

The agent's action space consists of four operations. \textbf{Propose} adds a new candidate key from KG hints or error analysis. \textbf{Profile} evaluates the candidate via SQL aggregation against temporally valid training data. \textbf{Score} ranks the candidate by validation accuracy, coverage, and minimum support. \textbf{Refine} updates the cascade by appending, reordering, or pruning levels. Throughout this paper, $\mathrm{MODE}(\cdot)$ denotes the most frequent value within each group. The key property is that the output of this loop is a deterministic lookup artifact, not a learned model. The LLM is used only during the offline discovery phase and is not needed at prediction time.

\smallskip\noindent\textbf{Candidate generation policy.}
Candidate keys are drawn from two sources. The first source is entities and attributes explicitly mentioned in the SALT-KG-provided description of field~$f$, such as the KG record that \textsc{ShippingPoint} depends on plant, shipping condition, and loading group. The second source is error-driven refinements, where the agent inspects which input features distinguish correctly predicted rows from errors and proposes additional key columns accordingly. Candidates are evaluated as single-entity keys, KG-suggested pairwise keys, or higher-order composites. The agent escalates to higher-order keys only when the accuracy gain exceeds a threshold~$\epsilon$.

\begin{algorithm}[t]
\SetAlgoLined
\KwIn{Target field $f$, KG descriptions $\mathcal{G}_f$, training tables $\mathcal{D}_{\text{train}}$, validation split $\mathcal{D}_{\text{val}}$, improvement threshold $\epsilon$, minimum support $m$, maximum depth $L_{\max}$}
\KwOut{Ordered cascade $(K_1, \ldots, K_L)$}
$\mathcal{H} \gets$ extract semantic hints from $\mathcal{G}_f$\;
$\mathcal{F} \gets$ initialize candidate frontier from $\mathcal{H}$\;
$\text{cascade} \gets [\;]$\;
\Repeat{\textnormal{no candidate improves validation accuracy by} $> \epsilon$ \textbf{or} $|\text{cascade}| \geq L_{\max}$}{
  \ForEach{candidate key $K \in \mathcal{F}$}{
    $\text{mapping}_K \gets \texttt{SQL: GROUP BY } K,f \texttt{; COUNT(*) and rank } f \texttt{ by frequency}$ on $\mathcal{D}_{\text{train}}$ where support $\geq m$\;
    $\text{acc}_K, \text{cov}_K \gets$ evaluate $\text{mapping}_K$ on $\mathcal{D}_{\text{val}}$\;
  }
  $K^* \gets \arg\max_K \text{acc}_K$ among candidates with $\text{cov}_K > 0$\;
  \If{$\text{acc}_{K^*} - \text{acc}_{\text{current}} > \epsilon$}{
    Append $K^*$ to cascade\;
    Analyze residual errors $\to$ update~$\mathcal{F}$ with error-driven candidates\;
  }
}
Append global target distribution as final level\;
\Return{cascade}
\caption{KG-Guided Cascade Discovery}
\label{alg:discovery}
\end{algorithm}

The loop terminates when (i)~no candidate improves validation accuracy by more than~$\epsilon$, (ii)~candidate support falls below~$m$, or (iii)~the maximum cascade depth~$L_{\max}$ is reached. In practice, cascades converge within 3 to 5 iterations per field.

\smallskip\noindent\textbf{Example trace for \textsc{CustomerPaymentTerms}.}
The SALT-KG-provided description of this field says that the system defaults to the key specified in the customer or vendor master record, which is the central ERP record for a business partner, while users can still enter a key during document entry. The agent interprets this as evidence that (1)~the customer identity is the primary determinant, and (2)~the document type may override the default. The following compressed trace shows one complete run of Algorithm~\ref{alg:discovery}.

\begin{enumerate}
    \item The KG hint points to defaults from the customer master. The agent queries \texttt{GROUP BY SoldToParty}. The result is 80.0\% accuracy with 10K keys, so the key is retained as fallback level~$L_1$.
    \item The KG hint indicates that document type may override the default. The agent queries \texttt{GROUP BY SoldToParty, SalesDocType}. The result is 82.9\% accuracy with 69K keys and 95.3\% coverage, so the key is promoted to~$L_0$.
    \item Error analysis suggests broader organizational keys. The agent queries \texttt{GROUP BY SalesOrg, DocType}. The gain is below~$\epsilon$, so the key is rejected.
    \item The coverage check shows that uncovered rows remain. The agent appends the global target distribution and finalizes the cascade.
\end{enumerate}

\noindent Without KG guidance, a minimal entity-level baseline would try only per-customer MODE, achieving 80.0\% (see ablation, Section~\ref{sec:ablation}). The KG hint about document-type override directly leads to the composite key that yields~$+2.9\%$.

\subsection{Cascade Codification (Build)}

A finalized cascade for field~$f$ is a sequence of levels $(K_1, \ldots, K_L)$, where each $K_l$ is a composite key. For each level we extract an empirical ranked list.
\[
\text{mapping}[l][k] = [(y_1,n_1),(y_2,n_2),\ldots], \quad n_1 \geq n_2 \geq \cdots,
\]
where $n_j$ is the training support of target value $y_j$ under key value~$k$, and the total support for~$k$ must exceed $\text{min\_support}_l$. The top-ranked value $y_1$ is the statistical mode and is used for top-1 accuracy, while the full ranked list is used for MRR. At prediction time, levels are queried in order, and the first match returns the ranked list for that level. This mirrors SAP's Access Sequence mechanism but is derived statistically rather than read from system configuration. For \textsc{CustomerPaymentTerms}, the cascade uses $L_0 = (\text{SoldToParty}, \text{SalesDocType})$, then $L_1 = (\text{SoldToParty})$, and finally $L_2 = \text{global distribution}$.

\subsection{Inference (Predict)}

The output of Phase~2 is a set of JSON lookup files. Prediction requires only key lookup, so no LLM, KG, or model inference is needed at runtime. Cascade sizes range from 88~keys (\textsc{Plant}) to 174K~keys (\textsc{HeaderIncoterms}).

\paragraph{Why cascades work in configured enterprise data.}
The method is effective when the target field has low conditional entropy given a small set of business keys. In such cases, historical transactions contain repeated evidence of the same configuration rule. A cascade lookup estimates this conditional distribution directly and preserves exact high-cardinality keys that embedding models may smooth away. The knowledge graph helps reduce the combinatorial search over column subsets by pointing to keys that have business meaning. The LLM does not replace statistical validation. It acts as a controller that translates semantic hints and residual errors into SQL candidates, while every accepted rule is selected by deterministic validation on held-out data.

\section{Experiments}
\label{sec:experiments}

\subsection{Setup}

We evaluate primarily on SALT-KG~\cite{mulang2025saltkg}, covering 8~autocomplete tasks across 4~SAP tables with temporal split at 2020-07-01 (train with 1.9M rows and test with 403K rows). For the generalization study (Section~\ref{sec:generalization}), we evaluate on two additional RelBench~v2~\cite{gu2026relbenchv2} datasets, namely rel-arxiv for academic papers with 53-class author-category prediction and rel-stack for StackOverflow with 3-class badge-type prediction. We compare against the baselines reported in SALT~\cite{klein2024salt} and SALT-KG~\cite{mulang2025saltkg}, and the heterogeneous GraphSAGE GNN from RelBench~v2~\cite{gu2026relbenchv2}. All cascade lookups use only temporally valid data before the evaluation split cutoff to prevent data leakage.

\paragraph{Metrics.}
Accuracy measures whether the top-ranked predicted value equals the held-out target value. MRR measures the reciprocal rank of the true value in the returned candidate list. For cascade methods, each matched lookup table stores target values ranked by their training frequency. The top-ranked value is the statistical mode and is used for accuracy, while the full ranked list is used for MRR. Thus, MODE is not an evaluation metric, but the estimator used to construct lookup entries from temporally valid training data.

\paragraph{Rule discovery implementation.}
The offline rule discovery agent (Section~\ref{sec:discovery}) uses Claude Opus~4.6 with temperature~0. Because every proposed candidate is evaluated deterministically by SQL on temporally valid data, the LLM affects the search process but not the scoring rule. To check sensitivity, we re-ran discovery on two representative fields (\textsc{CustomerPaymentTerms}, \textsc{Plant}) with GPT-5.3-Codex and obtained identical cascade structures and test accuracy. This does not constitute a full LLM-sensitivity study, but it suggests that the accepted rules are primarily constrained by the data and KG evidence.

\subsection{SALT-KG Results}

\begin{table}[t]
\centering
\caption{MRR on rel-salt. Top rows report baselines from SALT~\cite{klein2024salt} without KG. \textit{Best (w/ KG)} is the per-field maximum after adding KG-derived text embeddings~\cite{mulang2025saltkg}, computed from published base MRR + $\Delta$MRR at 2-decimal precision. Bottom row reports our cascade method.}
\label{tab:mrr}
\footnotesize
\resizebox{\textwidth}{!}{%
\begin{tabular}{lcccccccc|c}
\toprule
Method & PayTrm & Sales Grp. & Hdr Inc. & Item Inc. & ShipCond & ShipPt & Plant & Sales Off. & Avg \\
\midrule
Majority Class & 0.23 & 0.05 & 0.62 & 0.62 & 0.41 & 0.54 & 0.59 & 0.99 & 0.51 \\
XGBoost~\cite{chen2016xgboost} & 0.57 & 0.51 & 0.70 & 0.70 & 0.68 & 0.95 & 0.99 & 0.99 & 0.76 \\
LightGBM~\cite{ke2017lightgbm} & 0.10 & 0.02 & 0.73 & 0.73 & 0.51 & 0.28 & 0.61 & 0.99 & 0.50 \\
CatBoost~\cite{prokhorenkova2018catboost} & 0.44 & 0.16 & 0.81 & 0.80 & 0.71 & 0.80 & 0.99 & 0.99 & 0.71 \\
CARTE~\cite{kim2024carte} & 0.62 & 0.46 & 0.77 & 0.75 & 0.74 & 0.97 & 0.99 & 0.99 & 0.79 \\
AutoGluon~\cite{erickson2020autogluon} & 0.52 & 0.34 & 0.78 & 0.78 & 0.74 & 0.98 & 0.99 & 0.99 & 0.77 \\
GraphSAGE~\cite{hamilton2017graphsage} & 0.39 & 0.20 & 0.59 & 0.64 & 0.59 & 0.97 & 0.99 & 0.99 & 0.67 \\
\midrule
Best (w/ KG) & 0.65 & 0.51 & 0.81 & 0.80 & 0.78 & 0.98 & 0.99 & \textbf{1.00} & 0.815 \\
\midrule
Ours (cascade) & \textbf{0.852} & \textbf{0.760} & \textbf{0.840} & \textbf{0.840} & \textbf{0.798} & \textbf{0.986} & \textbf{0.993} & 0.997 & \textbf{0.883} \\
\bottomrule
\end{tabular}}
\end{table}

\begin{table}[t]
\centering
\caption{Accuracy comparison with RelBench~v2 baseline. GNN accuracy values are taken directly from the RelBench~v2 leaderboard~\cite{gu2026relbenchv2} using its default heterogeneous GraphSAGE~\cite{hamilton2017graphsage} architecture, hyperparameters, and training protocol. We did not re-train or tune the GNN.}
\label{tab:accuracy}
\begin{tabular}{lcc}
\toprule
Target Field & Ours (Acc) & GNN (Acc)\\
\midrule
\textsc{CustomerPaymentTerms} & \textbf{82.86\%} & 37.47\% \\
\textsc{SalesGroup} & \textbf{70.00\%} & 15.76\% \\
\textsc{HeaderIncoterms} & \textbf{77.22\%} & 62.23\% \\
\textsc{ItemIncoterms} & \textbf{77.15\%} & 69.36\% \\
\textsc{ShippingCondition} & \textbf{69.56\%} & 56.85\% \\
\textsc{ShippingPoint} & \textbf{98.65\%} & 98.39\% \\
\textsc{Plant} & \textbf{99.68\%} & 99.46\% \\
\textsc{SalesOffice} & \textbf{99.91\%} & 99.88\% \\
\bottomrule
\end{tabular}
\end{table}

Our method outperforms the evaluated GNN baseline on all 8~fields (Tables~\ref{tab:mrr} and~\ref{tab:accuracy}). The largest gains occur on high-cardinality, semantically complex fields (\textsc{SalesGroup}, \textsc{CustomerPaymentTerms}), where the GNN's message-passing mechanism does not capture well the business logic encoded in SAP's Condition Technique. For near-deterministic organizational mappings (\textsc{Plant}, \textsc{ShippingPoint}, \textsc{SalesOffice}), both methods converge above 98\%.

To connect accuracy to operational impact, consider \textsc{CustomerPaymentTerms}. The GNN assigns the correct payment term to only 37.5\% of orders, meaning that 62.5\% would require manual correction. The cascade achieves 82.9\%, reducing the manual correction rate to 17.1\%. For \textsc{SalesGroup} (commission-relevant), the GNN's 15.8\% accuracy is insufficient for operational use, while the cascade's 70.0\% makes it viable as a default suggestion.

\begin{figure}[t]
\centering
\includegraphics[width=\textwidth]{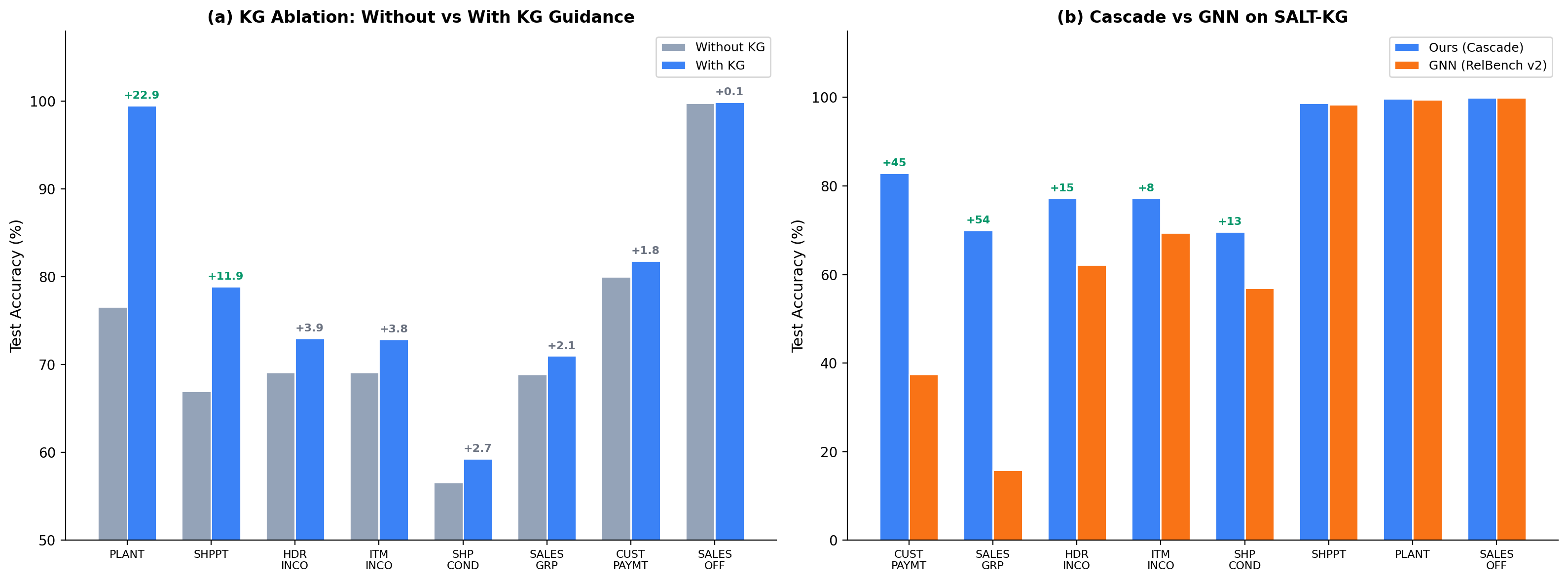}
\caption{(a) Semantic key-selection ablation before full error-driven refinement. (b) Final cascade vs.\ GNN using heterogeneous GraphSAGE from RelBench~v2~\cite{gu2026relbenchv2}, with per-field test accuracy on SALT-KG. Numbers above bars indicate absolute accuracy difference.}
\label{fig:results}
\end{figure}

\subsection{Semantic Key-Selection Ablation}
\label{sec:ablation}

To quantify the value of semantic key selection, we compare a schema-only entity MODE baseline with a restricted KG-guided cascade that uses the initial KG-suggested keys but not the full error-driven refinement and fallback optimization used in Tables~\ref{tab:mrr} and~\ref{tab:accuracy}. This ablation is intended to isolate whether KG-provided hints point the search toward more useful keys. It is not an exhaustive comparison against all possible rule-mining methods.

\begin{table}[t]
\centering
\caption{Ablation of semantic key selection on SALT-KG test set (accuracy). \textit{Restricted KG} denotes a KG-guided cascade before the full error-driven refinement and fallback optimization used in the main results.}
\label{tab:ablation}
\begin{tabular}{lccc}
\toprule
Field & Schema-only & Restricted KG & $\Delta$ \\
\midrule
\textsc{Plant} & 76.6\% & 99.5\% & \textbf{+22.9} \\
\textsc{ShippingPoint} & 67.0\% & 78.9\% & \textbf{+11.9} \\
\textsc{HeaderIncoterms} & 69.1\% & 73.0\% & +3.8 \\
\textsc{ItemIncoterms} & 69.1\% & 72.9\% & +3.8 \\
\textsc{ShippingCondition} & 56.6\% & 59.3\% & +2.6 \\
\textsc{SalesGroup} & 68.9\% & 71.0\% & +2.2 \\
\textsc{CustomerPaymentTerms} & 80.0\% & 81.8\% & +1.8 \\
\textsc{SalesOffice} & 99.8\% & 99.9\% & +0.1 \\
\midrule
\textbf{Average} & \textbf{74.1\%} & \textbf{79.5\%} & \textbf{+5.4} \\
\bottomrule
\end{tabular}
\end{table}

The value of semantic key selection varies by field type. Cross-entity fields (\textsc{Plant}, \textsc{ShippingPoint}) show the largest gains because the KG points to combinations of plants, shipping conditions, and loading groups, a relationship invisible from table schemas alone. Composite-key fields (\textsc{HeaderIncoterms}, \textsc{ItemIncoterms}, \textsc{ShippingCondition}, \textsc{SalesGroup}) gain between 2\% and 4\% because KG hints indicate which additional columns may disambiguate the target. Already-simple fields (\textsc{SalesOffice}, \textsc{CustomerPaymentTerms}) gain little because they are nearly determined by a single entity without KG guidance.

\subsection{Generalization Study}
\label{sec:generalization}

To test whether cascade lookups generalize beyond enterprise data, we apply the same cascade principle to two public RelBench~v2~\cite{gu2026relbenchv2} autocomplete tasks. These datasets do not provide expert enterprise KGs analogous to the SAP ontology, so these experiments use task-specific temporal and entity-level keys rather than the full KG-guided discovery loop. They therefore serve primarily to identify the boundary conditions under which cascades are and are not effective. All aggregation queries are temporally filtered to use only data before the evaluation cutoff, ensuring no label leakage.

\paragraph{rel-arxiv / author-category (53-class).}
For each author, we compute a temporal-weighted MODE(category) from their publications, giving higher weight to recent papers ($4{\times}/3{\times}/2{\times}/1{\times}$ for last 1/2/3/older years). This recency weighting improves test accuracy from 49.7\% (uniform MODE) to 50.3\%. On the validation set it achieves 52.8\%, slightly exceeding GNN (52.6\%). Citation- and coauthor-based fallbacks proved inferior to global mode due to topic breadth in cited papers and adjacent subfields among coauthors.

\paragraph{rel-stack / badges-class (3-class, gold/silver/bronze).}
After RelBench removes the target and metadata columns, badges contain only (Id, UserId, Date). The strongest cascade is per-user MODE(Class). Multi-hop enrichment through user activity (posts, votes, comments) was ineffective because bronze badges dominate the distribution so heavily that MODE returns bronze for every activity level, including the most active users.

\begin{table}[t]
\centering
\caption{Generalization study (accuracy on RelBench~v2 val and test sets).}
\label{tab:generalization}
\begin{tabular}{llccccc}
\toprule
Dataset & Task & Classes & Ours (Val) & GNN (Val) & Ours (Test) & GNN (Test) \\
\midrule
rel-arxiv & author-cat. & 53 & 52.8\% & 52.6\% & 50.3\% & 50.7\% \\
rel-stack & badges-cls. & 3 & 77.6\% & 80.0\% & 79.3\% & 82.8\% \\
\bottomrule
\end{tabular}
\end{table}

\paragraph{Why cascades match but do not surpass GNNs on non-enterprise data.}
Unlike SALT-KG's configured determinism, these tasks lack explicit business rules that cascade lookups can exploit.
\begin{enumerate}
    \item Cold-start entities dominate. On rel-arxiv, 29\% of test authors have no publications before the evaluation cutoff. The cascade can only predict the global mode for these authors (${\sim}2\%$ accuracy). GNNs face the same cold-start problem, explaining why both methods achieve only ${\sim}50\%$ accuracy.
    \item $\mathrm{MODE}(\cdot)$ is a lossy aggregation. On rel-stack, bronze badges constitute 86\% of training data, so $\mathrm{MODE}(\cdot)$ returns bronze for every feature combination. GNNs overcome this through learned aggregation functions that amplify minority-class signals, achieving 82.8\% vs.\ our 79.3\%.
    \item No configured determinism to recover. On rel-arxiv, an author's future category is correlated with past publications, not determined. On rel-stack, badge class is determined by threshold rules such as Great Answer requiring $\geq$100 upvotes. These rules require per-instance temporal reasoning, not entity-level aggregation.
\end{enumerate}

\subsection{Drift Root-Cause Analysis}
\label{sec:drift}

\begin{figure}[t]
\centering
\includegraphics[width=\linewidth]{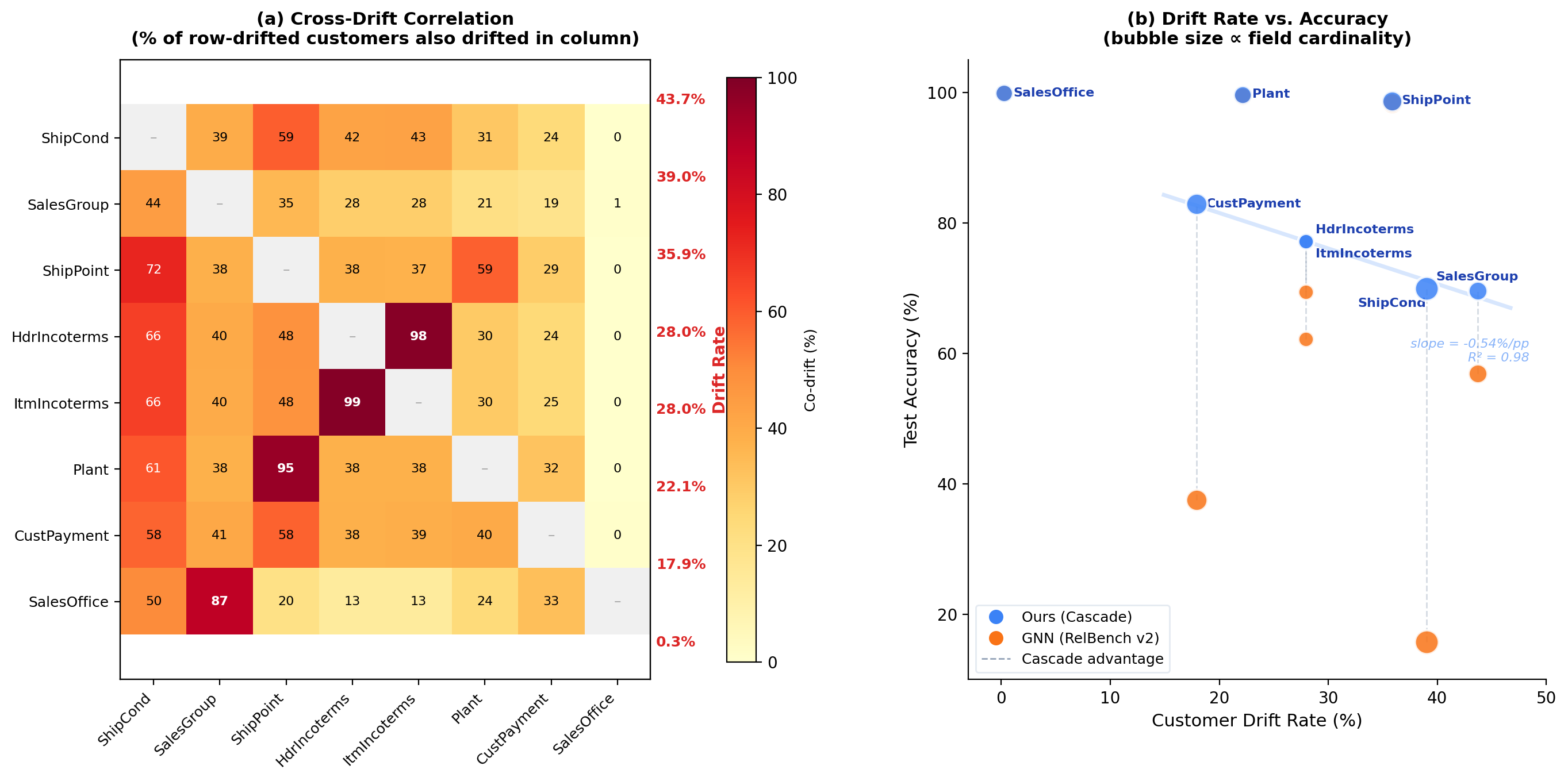}
\caption{Drift analysis. (a)~Cross-drift conditional correlation matrix, where asymmetric cells show $P(\text{column drifts} \mid \text{row drifts})$. (b)~Drift rate vs.\ test accuracy, with bubble size proportional to field cardinality.}
\label{fig:drift}
\end{figure}

The temporal split (train $<$ 2020-07-01 $<$ test) exposes concept drift, with per-customer MODE values changing between training and test periods. For each of the 5{,}576 customers present in both splits, we compute (1)~the per-field drift rate, defined as the fraction whose dominant value changed, and (2)~the cross-drift correlation, defined as the fraction of customers who drifted in field~$B$ among those who drifted in field~$A$. Figure~\ref{fig:drift} summarizes both analyses.

\paragraph{Cross-drift structure.}
Three structural patterns emerge.
\begin{enumerate}
    \item Coupled pairs move in lockstep. HdrIncoterms$\leftrightarrow$ItmIncoterms co-drift reaches between 98\% and 99\%, confirming they are governed by the same business decision. Plant$\to$ShipPoint co-drift reaches 95\%, so when the plant changes, the shipping point almost always follows.
    \item Upstream-downstream asymmetry reveals causal direction. Plant$\to$ShipPoint is 95\%, but ShipPoint$\to$Plant is only 58\%, which indicates that shipping point reassignment frequently occurs without changing the plant. Similarly, CustPayment$\to$ShipCond is 57\%, but ShipCond$\to$CustPayment is only 23\%.
    \item Stable fields are nearly drift-free. SalesOffice drifts in only 0.3\% of customers (14/5{,}576).
\end{enumerate}

\paragraph{Drift as a performance ceiling.}
Among the four non-trivial fields (accuracy $< 95\%$), drift rate and test accuracy are consistent with a linear anti-correlation (slope $= -0.54$ pp accuracy per pp drift, $R^2 = 0.98$). Because this fit is based on only four fields, the high $R^2$ should be interpreted as directionally suggestive rather than statistically conclusive. \textsc{CustomerPaymentTerms} has the lowest drift (17.8\%) and highest accuracy (82.9\%), while \textsc{ShippingCondition} has the highest drift (43.8\%) and lowest accuracy (69.6\%). This ceiling is method-agnostic because both cascades and GNNs face the same organizational changes.

For \textsc{HeaderIncoterms}, drifted customers show a directional DDP$\to$DAP transition (495 vs.\ 89 customers), consistent with documented supply chain risk reallocation during COVID-19 disruptions~\cite{davis2021incoterms}. For \textsc{SalesGroup}, 76\% of drifted customers retained the same Sales Area entirely, confirming that drift is driven by internal team reassignment rather than customer behavior changes.

\subsection{Error Analysis and Case Studies}
\label{sec:cases}

\paragraph{Success case with payment terms resolved by customer ${\times}$ document type.}
Consider a customer who places standard orders (document type TA). The cascade's finest level (SoldToParty $\times$ SalesDocumentType) maps this combination to payment term 03 (net 30 days), a mapping stable across the entire training period that correctly predicts all test-period orders. The cascade directly encodes the business rule that this customer's contract specifies net-30 for standard orders. Failure cases are structurally tied to drift. As shown in Section~\ref{sec:drift}, organizational restructuring (e.g., sales team reassignment) renders historical lookup rules obsolete in a way that no method can predict from transaction data alone.

\paragraph{Governance case with correlated drift revealing upstream configuration dependency.}
The cross-drift analysis (Figure~\ref{fig:drift}a) reveals that when a customer's \textsc{Plant} assignment changes, \textsc{ShippingPoint} co-drifts 95\% of the time, but the reverse holds only 58\%. This asymmetry tells configuration owners that plant reassignment should trigger shipping point review, but not vice versa. Similarly, co-drift between \textsc{HeaderIncoterms} and \textsc{ItemIncoterms} reaches between 98\% and 99\%, confirming they are governed by the same business decision, a validation that individual field-level audits might miss.

\section{Discussion}
\label{sec:discussion}

\paragraph{When to use cascades in practice.}
The central lesson from our three-domain evaluation is that the quality of available domain knowledge matters more than the sophistication of the model. The differentiator is the alignment between domain knowledge and observable data. Cascades excel when a domain-expert KG provides determination hints that correspond to available data columns, match GNNs when no such expert KG or stable rule structure exists (rel-arxiv with 50.3\% vs.\ 50.7\%), and underperform when prediction requires per-instance temporal reasoning that GNNs learn implicitly (rel-stack with 79.3\% vs.\ 82.8\%). Practitioners should prefer GNNs when discriminative signal resides in graph neighborhoods or temporal patterns, and use drift analysis as an operational diagnostic regardless of method, since the observed relationship between drift rate and accuracy (Section~\ref{sec:drift}) enables proactive identification of entities requiring configuration review. In this sense, the method recovers part of the declarative rule structure visible in historical transactions, while procedural configuration changes and organizational restructurings remain outside the observable data.

\paragraph{Practical advantages.}
Beyond accuracy, cascades offer interpretability (each prediction traces to a specific composite key), no model-serving cost at inference time (JSON key lookup on CPU, with 12~MB total storage for 8~fields vs.\ $\sim$200~MB model weights), and rapid updates (rebuild in minutes vs.\ hours of GPU retraining). We recommend field-specific deployment modes, including editable suggestions for all fields, conservative auto-fill for near-deterministic fields ($>$99\% accuracy), and audit-only logging for high-drift fields.

\paragraph{Prototype-to-production path.}
The current evaluation is offline. A production deployment should first run in shadow mode, where cascade suggestions are logged but not automatically applied. Operational metrics should include suggestion acceptance rate, operator override rate, post-entry correction rate, lookup latency under concurrent load, and the precision of drift alerts as judged by configuration owners. These measurements are required before claiming realized workload reduction in production.

\paragraph{Reproducibility.}
All code, cascade JSON files, evaluation scripts, and the drift analysis toolkit are publicly released. Experiments are reproducible from the released code and public SALT-KG data~\cite{mulang2025saltkg}.

\paragraph{Ethics and governance.}
All experiments use pseudonymized keys (integer IDs) without access to personally identifiable information. Cascade mappings should be audited for correlations with protected attributes before deployment. Cascades require periodic rebuilding (e.g., quarterly), with drift alerts flagging customers exceeding historical change rates. Predictions are surfaced as editable suggestions rather than autonomous overwrites, and operators retain full override authority. Each prediction logs the matched cascade level and composite key, supporting post-hoc analysis and regulatory compliance.

\paragraph{Threats to validity.}
SALT-KG represents a single SAP S/4HANA ERP system instance. The generalization study partially addresses this, but neither public dataset has comparable configuration richness. The public datasets do not contain expert-curated determination KGs comparable to the enterprise setting, so the generalization study evaluates the cascade principle rather than the full KG-guided workflow. GNN baselines use RelBench~v2's default protocol without task-specific tuning. The LLM sensitivity check covers only two representative fields, and weaker models may fail to translate KG hints effectively. The temporal split at 2020-07-01 coincides with COVID-19, and observed drift patterns may reflect pandemic-specific disruptions.

\paragraph{Limitations.}
(1)~Cascades require configured determinism to outperform GNNs. On general relational data they match but do not exceed GNN performance.
(2)~Cold-start entities fall through to coarse defaults with no viable relational fallback. In deployment, predictions that reach only coarse organizational fallbacks or the global mode should be surfaced as low-confidence suggestions rather than auto-filled values.
(3)~The system has been validated offline on historical data and has not yet been deployed in a production ERP environment.
(4)~We did not include classical rule-mining baselines such as association rules, FP-Growth, RuleFit, or decision lists. This limits our ability to separate the benefit of KG-guided key selection from the benefit of symbolic rule extraction in general.

\paragraph{Future directions.}
Two extensions are most promising. The first is temporal cascades that perform per-instance temporal-windowed aggregation, enabling rule reconstruction for threshold-based systems like StackOverflow badges. The second is hybrid cascade-GNN architectures that use cascade predictions as features or priors for GNN training, potentially combining interpretability with learned generalization for unseen or recently reorganized entities.

\section{Conclusion}
\label{sec:conclusion}

The results show that KG-grounded cascade lookups, produced by LLM-guided SQL profiling and deployed as lightweight JSON files, outperform evaluated GNN baselines on all 8~SALT-KG enterprise fields. A generalization study on two public datasets reveals clear boundary conditions. Cascades match GNNs when field values are governed by stable, recoverable rules, but underperform when prediction requires temporal graph reasoning. The drift analysis demonstrates that many residual errors are associated with organizational changes, providing configuration owners with actionable diagnostics. For practitioners, the results suggest that investing in domain knowledge assets can be more valuable than adopting more complex models when the underlying data is governed by recoverable configuration rules.

\subsubsection*{Acknowledgment}
The author was funded by the InsightsNet project (funded by the Federal Ministry of Education and Research (BMBF) under grant no. 01UG2130A). 
I gratefully acknowledge the Research group Corpus and Computational Linguistics, Institute of Linguistics and Literary Studies, TU Darmstadt.
I gratefully acknowledge support from the hessian.AI Service Center (funded by the Federal Ministry of Research, Technology and Space, BMFTR, grant no. 16IS22091) and the hessian.AI Innovation Lab (funded by the Hessian Ministry for Digital Strategy and Innovation, grant no. S-DIW04/0013/003).
I gratefully acknowledge the computing time granted by the Resource Allocation Board and provided on the supercomputer Emmy/Grete at NHR-Nord@Göttingen as part of the NHR infrastructure. The calculations for this research were conducted with computing resources under the project \verb|nhr_he_starter_25563|.

\subsubsection*{AI Usage.}
LLMs were used as part of the proposed offline rule-discovery workflow, as described in Section~\ref{sec:discovery}. For manuscript preparation, LLMs were used only for readability and grammar assistance. The author reviewed all content and takes full responsibility.

\subsubsection*{Disclosure of Interests.}
The author has no competing interests to declare that are relevant to the content of this article.
\bibliographystyle{splncs04}
\bibliography{custom}

\end{document}